\documentclass[12pt]{amsart}
\usepackage[english]{babel}
\usepackage{egothic}
\usepackage[T1]{fontenc}

\usepackage{amsmath}
\usepackage{amsthm}
\usepackage{amssymb}
\usepackage{amsfonts} 
\usepackage{mathrsfs} 
\usepackage{pb-diagram} 
\usepackage[all]{xy}
\usepackage{mathrsfs}
\usepackage{enumerate}
\usepackage{physics}
\usepackage[notcite, final, notref]{showkeys}
\usepackage{dsfont}
\usepackage[dvips]{color}
\newtheorem{theorem}{Theorem}[section]

\newtheorem{corollary}[theorem]{Corollary}
\newtheorem{definition}[theorem]{Definition}

\theoremstyle{definition}
\newtheorem{remark}[theorem]{Remark}

\newcommand{\BC}{{\mathbb B}{\mathbb C}}
\newcommand{\D}{{\mathbb D}}

\renewcommand{\i}{{\bf i}}
\renewcommand{\j}{{\bf j}}
\renewcommand{\k}{{\bf k}}
\newcommand{\C}{{\mathbb C}}
\newcommand{\R}{{\mathbb R}}
\newcommand{\e}{{\bf e}}

\renewcommand{\Re}{\mathrm{Re}}

\usepackage{xcolor}

\begin{document}

\title[Bicomplex Neural Networks]
{A note on the complex and bicomplex valued neural networks}

\author[D. Alpay]{Daniel Alpay}
\address{(DA) Schmid College of Science and Technology \\
Chapman University\\
One University Drive
Orange, California 92866\\
USA}
\email{alpay@chapman.edu}

\author[K. Diki]{Kamal Diki}
\address{(KD) Schmid College of Science and Technology \\
Chapman University\\
One University Drive
Orange, California 92866\\
USA}
\email{diki@chapman.edu}

\author[M. Vajiac]{Mihaela Vajiac}
\address{(MV) Schmid College of Science and Technology \\
Chapman University\\
One University Drive
Orange, California 92866\\
USA}
\email{mbvajiac@chapman.edu}

\thanks{D. Alpay thanks the Foster G. and Mary McGaw Professorship in
  Mathematical Sciences, which supported his research}

\begin{abstract}
 In this paper we first write a proof of the perceptron convergence algorithm for the complex multivalued neural networks (CMVNNs). Our primary goal is to formulate and prove the perceptron convergence algorithm for the bicomplex multivalued neural networks (BMVNNs) and other important results in the theory of neural networks based on a bicomplex algebra. 
 \end{abstract}

\maketitle

\noindent {\em }
\date{today}


\section{Introduction}
\setcounter{equation}{0}

The perceptron was created by a psychologist (Frank Rosenblatt, circa 1958), based on the model of the neuron developed by McCulloch and Pitts \cite{mcpitts}, and with the aim to define the Hebb model of learning (see \cite{hebb1949organisation} or, more precisely \cite[p. 514]{block1962perceptron}). The perceptron is a linear classifier, with a non-linear activation function, and an important example of a supervised learning algorithm. 

Note that related works in theoretical mathematics, (in the setting of $\mathbb R^N$, see \cite[p. 93]{nilsson1965learning}), were also done earlier in 1954 by Agmon (see \cite{agmon}), and by Motzkin and Schoenberg (see \cite{MR62787}).  A review of proofs and applications of the perceptron theorem in the real setting may be found in \cite[pp. 92-93]{nilsson1965learning} and we mention in particular the classical works of~\cite{block1962perceptron,minskypapert,zbMATH03189712,MR0135635,singleton1962test}, as well as more recent ones~\cite{abbn, TM}. We also give a quick reminder of these concepts in Subsection~\ref{classical}.\\

The perceptron is also the smallest unit capable of learning a classification problem between two classes and its ability to do so is proven in the perceptron convergence theorem, a recursive procedure which allows one to find a separating hyperplane for a separable family of vectors in $\mathbb R^N$. In this present work we extend this important result and procedure to the bicomplex setting, which we introduce in Section~\ref{bicomplex}. \\

There has been a revived interest in the subject of perceptrons due to the the needs of quantum computing and we refer the reader to~\cite{SSP, LWYZ}, where a quantum perceptron model is introduced. We hope that our present work will open new avenues in this direction as well.\\

We will first extend the perceptron convergence algorithm to the complex case and re-write a proof here. We then set up the basis of the perceptron algorithm in the bicomplex case and prove its convergence. A proof of the convergence algorithm in the complex case exists in~\cite{georgiou} (as well as in~\cite{aizenbergtwice}). However, for completeness, we include our own proof in this case, as it sheds light on the bicomplex perceptron theorem, which is the main aim of our paper.\\

Just as in the case of complex numbers, where the individual components of the complex number can be treated independently as two real numbers, the bicomplex space can be treated as two complex or four real numbers. In  \cite{Hirose-Yoshida}, it has been shown that the operation of complex multiplication limits the degree of freedom of the complex valued neural network at the synaptic weighting, therefore a CVNN is not quite equivalent to a two-dimensional real-valued neural network. 
In the same way a BVNN is not quite equivalent to a four-dimensional real-valued neural network, due to the bicomplex multiplication rules.
This being said, complex-valued neural network research in signal processing applications include channel equalization \cite{Huang, Sol2002}, satellite communication equalization \cite{BMPU} and, from a biological perspective, the complex-valued representation has been used in~\cite{RS2014}.\\

We start with short review of the regular perceptron algorithm in the real case, then set up the complex perceptron environment. The bicomplex approach gives a better way of combining two-valued complex networks compared to the usual two-valued approach in the literature~\cite{aizenbergtwice}, as we can treat the theory as a single variable one due to its algebra structure. In our approach the structure of the bicomplex algebra is essential in providing a convergent bicomplex perceptron algorithm as seen in Section~\ref{BC_perceptron}.

This work is part of a general hypercomplex setting and we refer the reader to~\cite{BS2008} and~\cite{} for other examples, such as the Clifford algebra.

\subsection{Preliminaries: Classical Perceptron Algorithm}
\label{classical}
The setting is a (possibly infinite) family $\mathcal C$  of vectors in the feature space $\mathbb R^N$ (for some fixed $N\in\mathbb N$), strictly divided into two classes
  $\mathcal C_+$ and $\mathcal C_-$ via an hyperplane: more precisely, we assume that there exists a unit norm vector $a\in\mathbb R^N$ such that
\begin{equation*}
  \label{c1c2}
x\in\mathcal C_+\,\,\iff\, a^t x>0\quad {\rm and}\quad x\in\mathcal C_-\,\,\iff\, a^tx<0,
\end{equation*}
and
\begin{equation*}
  \label{delta3456}
\exists \delta>0\,\, {\rm such \,\, that\,\,} \inf_{x\in\mathcal C}|a^tx|\ge \delta.
\end{equation*}
Under these hypothesis, $0\not \in \mathcal C$, and the hyperplane will not be unique. This condition is called {\em separability}.\\

The perceptron convergence theorem gives an iterative way to compute an hyperplane which separates the two classes after a {\bf finite} number of steps.
The coefficients of the equation of an hyperplane solving the problem (the weights) are learned via an algorithm, which is the content of the perceptron convergence theorem. A proof of this algorithm can be found in~\cite{haykin}.

\begin{theorem} Under the separability hypothesis, let $\mathcal C$ be a possibly not countable family of non-zero vectors in $\mathbb R^N$,  and let
  $x^{(1)},x^{(2)},\ldots$ be a countable family of elements of $\mathcal C$, each element often appearing infinitely many times. Then the sequence defined by $a^{(0)}=x^{(1)}$ and
  \begin{equation*}
    a^{(n+1)}=\begin{cases}\,\, a^{(n)}\,\,\hspace{15mm}{if}\,\,\ x^{(n)}\in\mathcal C_+ \,\,{and}\,\, (a^{(n)})^tx^{(n)}>0,\\
      \,\,a^{(n)}\,\,\hspace{15mm}{ if}\,\,\ x^{(n)}\in\mathcal C_- \,\,{ and}\,\,( a^{(n)})^tx^{(n)}<0,\\
      \,\,a^{(n)}+\dfrac{x^{(n)}}{\|x^{(n)}\|}\,\,\,{if}\,\,\ x^{(n)}\in\mathcal C_+ \,\,{ and}\,\, (a^{(n)})^tx^{(n)}\le 0,\\
      \,\,a^{(n)}-\dfrac{x^{(n)}}{\|x^{(n)}\|}\,\,\,{if}\,\,\ x^{(n)}\in\mathcal C_- \,\,{ and}\,\, (a^{(n)})^tx^{(n)}\ge 0,
    \end{cases}
    \label{273}
  \end{equation*}
  is stationary after a finite number $M$  of steps, (i.e. $a^{(n)}=a^{(M)}$ for any $n\leq M$), with
$\displaystyle M+1\le \frac{1}{\delta^2}$.
    \end{theorem}

    We denote by $M$ the number of times the sequence $a^{(0)},a^{(1)},\ldots$ changes
    from one index to the next one. Recall also that one assumes that there exists $a\in\mathbb R^N$ (but, as already mentioned, $a$ is not unique, and unknown) which answers the problem.
    Following the book~\cite{minskypapert}, the algorithm can be divided into four steps and the convergence follows.\\
 %
%
%

\begin{remark}The proof holds also in the case of a finite sequence. The result is then of interest when $\frac{1}{\delta^2}$ is much smaller than the cardinal of $\mathcal C$. The proof also
holds when the vectors belong to a general Hilbert space.
\end{remark}

   


The proof of the convergence algorithm relies heavily on the Cauchy-Schwarz Inequality, as well as the triangle inequality for the Euclidean metric on $\mathbb R^N$.
Variations of the algorithms are possible; see for instance \cite[p. 180]{duda}.

\begin{equation*}
  a^{(n+1)}=\begin{cases}\,\, a^{(n)}\,\,\hspace{15mm}{if}\,\,\ x^{(n)}\in\mathcal C_+ \,\,{and}\,\, (a^{(n)})^tx^{(n)}>0,\\
      \,\,a^{(n)}\,\,\hspace{15mm}{ if}\,\,\ x^{(n)}\in\mathcal C_- \,\,{ and}\,\,( a^{(n)})^tx^{(n)}<0,\\
      \,\,a^{(n)}+\e(n)\dfrac{x^{(n)}}{\|x^{(n)}\|}\,\,\,{if}\,\,\ x^{(n)}\in\mathcal C_+ \,\,{ and}\,\, (a^{(n)})^tx^{(n)}\le 0,\\
      \,\,a^{(n)}-\e(n)\dfrac{x^{(n)}}{\|x^{(n)}\|}\,\,\,{if}\,\,\ x^{(n)}\in\mathcal C_- \,\,{ and}\,\, (a^{(n)})^tx^{(n)}\ge 0,
    \end{cases}
  \end{equation*}
  where $\e(n)>0$. The sequence is stationary after a finite number of iterations when the weights sequence satisfies
  \[
    \lim_{M\rightarrow\infty}\frac{\sum_{m=1}^M\e(m)^2}{\left(\sum_{m=1}^M\e(m)\right)^2}=0.
      \]


In our paper we will generalize the classical perceptron algorithm to the set of bicomplex numbers and prove its convergence.

\subsection{Overview of our results}
We start with a short review of the bicomplex algebra in Section~\ref{bicomplex} where we re-introduce a hyperbolic valued modulus that will be used in the proof of the bicomplex algorithm. 
In Section~\ref{complex_Perc} we re-write a complete proof of the convergence of the perceptron algorithm in the complex case. 
The last part of the paper, in the section~\ref{BC_perceptron} is dedicated to the proof of analogues of the complex perceptron theorems in the bicomplex case.

\section{Introduction to Bicomplex Numbers}
\label{bicomplex}

The algebra of bicomplex numbers was first introduced by Segre in~\cite{Segre}. During the past decades, a few isolated works analyzed either the properties of bicomplex numbers, or the properties of holomorphic functions defined on bicomplex numbers, and, without pretense of completeness, we direct the attention of the reader first to the to book of Price,~\cite{Price}, where a full foundation of the theory of multicomplex numbers was given, then to some of the works describing some analytic properties of functions in the field~\cite{alss, CSVV, DSVV, mltcplx}. Applications of bicomplex (and other hypercomplex) numbers can be also found in the works of Alfsmann, Sangwine, Gl\"{o}cker, and Ell~\cite{AG, AGES}.\\

We now introduce, in the same fashion as~\cite{CSVV,bcbook,Price}, the key definitions and results for the case of holomorphic functions of complex variables. The algebra of bicomplex numbers is generated by two commuting imaginary units $\i$
and $\j$ and we will denote the bicomplex space by $\BC$.  The product of the two commuting units  $\i$ and $\j$ is denoted by $ k := \i\j$ and we note that $\k$ is a hyperbolic unit, i.e. it is a unit which squares to $1$.  Because of these various units in $\BC$, there are several
different conjugations that can be defined naturally. We will make use of these appropriate conjugations in this paper, and we refer the reader to~\cite{bcbook,mltcplx} for more information on bicomplex and multicomplex analysis.

\medskip
\subsection{Properties of the bicomplex algebra}
\label{bc}
The bicomplex space, $\BC$, is not a division algebra, and it has two distinguished zero
divisors, $\e_1$ and $\e_2$, which are idempotent, linearly independent
over the reals, and mutually annihilating with respect to the
bicomplex multiplication:
\begin{align*}
  \e_1&:=\frac{1+\k}{2}\,,\qquad \e_2:=\frac{1-\k}{2}\,,\\
  \e_1 \cdot \e_2 &= 0,\qquad
  \e_1^2=\e_1 , \qquad \e_2^2 =\e_2\,,\\
  \e_1 +\e_2 &=1, \qquad \e_1 -\e_2 = \k\,.
\end{align*}
Just like $\{1,\mathbf{j} \},$ they form a basis of the complex algebra
$\BC$, which is called the {\em idempotent basis}. If we define the
following complex variables in $\C(\i)$:
\begin{align*}
  \beta_1 := z_1-\i z_2,\qquad \beta_2 := z_1+\i z_2\,,
\end{align*}
the $\C(\i)$--{\em idempotent representation} for $Z=z_1+\j z_2$ is
given by
\begin{align*}
  Z &= \beta_1\e_1+\beta_2\e_2\,.
\end{align*}

The $\C(\i)$--idempotent is the only representation for which
multiplication is component-wise, as shown in the next lemma.

\begin{remark}
  \label{prop:idempotent}
  The addition and multiplication of bicomplex numbers can be realized
  component-wise in the idempotent representation above. Specifically,
  if $Z= a_1\,\e_2 + a_2\,\e_2$ and $W= b_1\,\e_1 + b_2\,\e_2 $ are two
  bicomplex numbers, where $a_1,a_2,b_1,b_2\in\C(\i)$, then
  \begin{eqnarray*}
    Z+W &=& (a_1+b_1)\,\e_1  + (a_2+b_2)\,\e_2   ,  \\
    Z\cdot W &=& (a_1b_1)\,\e_1  + (a_2b_2)\,\e_2   ,  \\
    Z^n &=& a_1^n \,\e_1  + a_2^n \,\e_2  .
  \end{eqnarray*}
  Moreover, the inverse of an invertible bicomplex number
  $Z=a_1\e_1 + a_2\e_2 $ (in this case $a_1 \cdot a_2 \neq 0$) is given
  by
  $$
  Z^{-1}= a_1^{-1}\e_1 + a_2^{-1}\,\e_2 ,
  $$
  where $a_1^{-1}$ and $a_2^{-1}$ are the complex multiplicative
  inverses of $a_1$ and $a_2$, respectively.
\end{remark}

One can see this also by computing directly which product on the
bicomplex numbers of the form
\begin{align*}
  x_1 + \i x_2 + \j x_3 + \k x_4,\qquad x_1,x_2,x_3,x_4\in\R
\end{align*}
is component wise, and one finds that the only one with this property
is given by the mapping:
\begin{align}
  \label{shakira}
  x_1 + \i x_2 + \j x_3 + \k x_4 \mapsto ((x_1 + x_4) + \i (x_2-x_3), (x_1-x_4) + \i (x_2+x_3))\,,
\end{align}
which corresponds exactly with the idempotent decomposition
\begin{align*}
  Z = z_1 + \j z_2 = (z_1- \i z_2)\e_1 + (z_1+ \i z_2)\e_2\,,
\end{align*}
where $z_1 = x_1+ \i x_2$ and $z_2 = x_3+ \i x_4$.

\begin{remark}
These split the bicomplex space in $\BC=\mathbb C \mathbf{e}_1\bigoplus \mathbb C \mathbf{e}_2$, as:
\begin{equation}
  Z=z_1+\j z_2=(z_1-\i z_2)\mathbf{e}_1+(z_1+\i z_2)\mathbf{e}_2=\lambda_1\e_1+\lambda_2\e_2.
\end{equation}
\end{remark}

Simple algebra yields:
\begin{equation}
\begin{split}
  z_1&=\frac{\lambda_1+\lambda_2}{2}\\
    z_2&=\frac{\i(\lambda_1-\lambda_2)}{2}.
  \end{split}
  \end{equation}

Because of these various units in $\BC$, there are several
different conjugations that can be defined naturally and we will now define the conjugates in the bicomplex setting, as in~\cite{CSVV,bcbook}

\begin{definition} For any $Z\in \BC$ we have the following three conjugates:
  \begin{eqnarray}
  \label{conj}
    \overline{Z}=\overline{z_1}+\j\overline{z_2}\\
     Z^{\dagger}=z_1-\j z_2\\
      Z^*=\overline{Z^{\dagger}}=\overline{z_1}-\j\overline{z_2}.
  \end{eqnarray}
\end{definition}

We refer the reader to~\cite{bcbook} for more details.

\bigskip

\subsection{Hyperbolic subalgebra and the hyperbolic-valued modulus}

A special subalgebra of $\BC$ is the set of hyperbolic numbers, denoted by $\D$.  This
algebra and the analysis of hyperbolic numbers have been studied, for
example, in~\cite{alss,bcbook,sobczyk} and we summarize
below only the notions relevant for our results.  A {\em hyperbolic number}
can be defined independently of $\BC$, by $\mathfrak{z}=x + \k y$,
with $x,y,\in\R$, $\k\not\in\R, \k^2=1$, and we denote by $\D$ the
algebra of hyperbolic numbers with the usual component--wise addition
and multiplication.  The hyperbolic {\em conjugate} of $\mathfrak z$
is defined by $\mathfrak{z}^\diamond := x - \k y$, and note that:
\begin{equation}
  \mathfrak{z}\cdot \mathfrak{z}^\diamond=x^2-y^2\in\R\,,
\end{equation}
which yields the notion of the square of the {\em modulus} of a
hyperbolic number $\mathfrak{z}$, defined by
$ |\mathfrak{z}|_{\D}^2:=\mathfrak{z}\cdot \mathfrak{z}^\diamond$.
\begin{remark}
It is worth noting that both $\overline{Z}$ and $Z^{\dagger}$ reduce to  $\mathfrak{z}^\diamond$ when $Z= \mathfrak{z}$. In particular $\e_2=\e_1^\diamond=\e_1^*=\e_1^{\dagger}$.
\end{remark}
Similar to the bicomplex case, hyperbolic numbers have a unique
idempotent representation with real coefficients:
\begin{align}
  \label{D_idempotent}
  \mathfrak{z}=s \e_1 + t \e_2 \,,
\end{align}
where, just as in the bicomplex case, $\displaystyle \e_1 = \frac{1}{2} \left( 1 + \k \right) $,
$\displaystyle \e_2= \frac{1}{2} \left( 1 - \k \right)$, and
$s:=x+y$ and $t:=x-y$. Note that $\e_1^\diamond=\e_2$ if we consider
$\D$ as a subset of $\BC$, as briefly explained in the remark above. We also observe that
$$
|\mathfrak{z}|_{\D}^2 = x^2 - y^2 = (x+y)(x-y) = st.
$$

The hyperbolic algebra $\D$ is a subalgebra of the bicomplex numbers
$\BC$ (see~\cite{bcbook} for details). Actually $\BC$ is the
algebraic closure of $\D$, and it can also be seen as the
complexification of $\D$ by using either of the imaginary unit $\i$ or
the unit $\j$.
\begin{definition}
Define the set $\D^+$ of {\em non--negative hyperbolic numbers} by:
\begin{align*}
  \D^+ &= \left\{ x + \k y \, \big| \, x^2 - y^2 \geq 0,  x \geq 0 \right\}
       = \left\{ x + \k y \, \big| \, x \geq 0,  | y | \leq x \right\} \\
       &= \{ s \e_1 + t \e_2 \, \big| \, s, t \geq 0 \}.
\end{align*}
\end{definition}
\begin{remark}
As studied extensively in~\cite{alss}, one can define a partial order
relation defined on $\D$ by:
\begin{align}
  \label{po}
  \mathfrak{z}_1 \preceq \mathfrak{z}_2\qquad\text{if and only if}\qquad
  \mathfrak{z}_2-\mathfrak{z}_1\in\D^+,
\end{align}
and we will use this partial order to study the
{\em hyperbolic--valued} norm, which was first introduced and studied
in~\cite{alss}.
\end{remark}

The Euclidean norm $\|Z\|$ on $\BC$, when it is seen as
$\C^2(\i), \C^2(\j)$ or $\R^4$ is:
\begin{align*}
  \|Z\| = \sqrt{ | z_1 | ^2 + | z_2 |^2 \, } = \sqrt{ \Re\left( | Z |_\k^2 \right) \, } = \sqrt{
  \, x_1^2 + y_1^2 + x_2^2 + y_2^2 \, }.
\end{align*}
As studied in detail in~\cite{bcbook}, in idempotent
coordinates $Z=\lambda_1\e_1+\lambda_2\e_2$, the Euclidean norm becomes:
\begin{align}
  \label{Euclidean_idempotent}
  \|Z\| = \frac{1}{\sqrt2}\sqrt{|\lambda_1|^2 + |\lambda_2|^2}.
\end{align}


It is easy to prove that
\begin{align}
  \|Z \cdot W\|  \leq  \sqrt{2} \left(\|Z\| \cdot  \|W\| \right),
\end{align}
and we note that this inequality is sharp since if $Z = W= \e$, one
has:
\begin{align*}
  \|\e_1 \cdot \e_1\| = \|\e_1\| = \frac{1}{\sqrt{ 2}} =  \sqrt{2}\, \|\e_1\| \cdot \|\e_1\|,
\end{align*}
and similarly for $\e_2$.

\begin{definition}
One can define a
{\em hyperbolic-valued} norm for $Z=z_1+\j z_2 = \lambda_1 \e_1+\lambda_2\e_2$
by:
\begin{align*}
  \| Z\|_{\D_+} := |\lambda_1|\e_1 + |\lambda_1|\e_2 \in\D^+.
\end{align*}
\end{definition}
It is shown in~\cite{alss} that this definition obeys the corresponding properties
of a norm, i.e. $ \| Z\|_{\D_+}=0$ if and only if $ Z=0$,
it is multiplicative, and it
respects the triangle inequality with respect to the order introduced
above. 

\subsection{Hyperbolic-valued modulus of vectors in $\BC$}
\label{BCvectors_norm}

The previous norm can be generalized to the space of $\BC$ vectors, i.e. elements of $\BC^n$, and we will also define an inner product on the space of vectors in $\BC$.
Let  $\displaystyle \langle X,\, Y \rangle$ be the usual Hermitian inner product on $\C^n$, then we have the following:

\begin{definition}
For any $\mathsf  X,\mathsf Y\in \BC^n$, we have the following $\D-$valued inner product
\begin{equation}
\label{ip_def_bc_n}
 \langle \mathsf  X,\, \mathsf  Y \rangle_{\D}= \langle X_1,\, Y_1 \rangle \e_1+ \langle X_2,\, Y_2 \rangle \e_2,
\end{equation}
where $\mathsf  X=X_1\e_1 +X_2\e_2$ and  $\mathsf  Y=Y_1\e_1 +Y_2\e_2$, and $X_l,\, Y_l\in \C^n$ for $l=1,2$.
\end{definition}

This inner product yields the hyperbolic-valued modulus of a vector  $\mathsf  X=X_1\e_1 +X_2\e_2$ as: $$\| \mathsf  X\|_{\D_+}=\|X_1\|\e_1 +\|X_2\|\e_2.$$

 \section{The convergence of the perceptron algorithm in the complex case } 
 \label{complex_Perc}
\setcounter{equation}{0}

In this section we will give a proof of the complex perceptron algorithm and pave the way to the bicomplex case. One of the first works on complex activation function was done by Naum Aizenberg \cite{AIPK1973, AZivko}, where a multi-valued neuron (MVN) is a neural element with $n$ complex inputs and one complex output on the unit circle, with complex-valued weights.
\begin{definition}\label{ActFunComplex}
Let $\displaystyle \varepsilon=\exp(\frac{2\pi \i}{k})$ be the the root of unity of order $k$. 
For $T\subset \mathbb{C}^n$ one can define the following activation function $P$ by $\displaystyle P(z)=\varepsilon^l$, whenever $\displaystyle\frac{2\pi \i l}{k}\leq Arg(z)<\frac{2\pi \i (l+1)}{k}$.
\end{definition}

Using this activation function, following~\cite{aizenbergtwice}, one can define the concept of a threshold function:

\begin{definition}
Let $n\geq 1$ and $T\subset \mathbb{C}^n$. Then, a complex valued function $f:T\longrightarrow \mathbb{C}$ is called a {\em threshold function} if there exists a weighting vector $W=(w_0, w_1,...,w_n)\in \mathbb{C}^{n+1}$ such that:
\begin{equation}
f(x_1,...,x_n)=P(w_0+\sum_{\ell=1}^{n}w_{\ell}x_{\ell}), \quad \text{ for any }
(x_1,...,x_n)\in T.
\end{equation}

\end{definition}

The activation function divides the complex plain into $k$ equal sectors, and implements a mapping of the entire complex plane onto the unit circle.
Here, we give a different proof of Theorem 2.1 in~\cite{aizenbergson}:
\begin{theorem}
Let $T$ be a bounded domain of $\mathbb{C}^n$, $f:T\longrightarrow \mathbb{C}$ a threshold function and $(w_0,0,...,0)$ a weighting vector of $f(x_1,...,x_n)$. Then, there exists $w'_0\in\mathbb{C}$ and $\delta>0$ such that $(w'_0,w_1,...,w_n)$ is a weighting vector of $f$ for every $w_1,...,w_n$ that satisfy $|w_j|<\delta$ with $j=1,...,n$.
\end{theorem}

\begin{proof}
In the $k-$valued case we have that $\displaystyle \varepsilon=e^{i\frac{2\pi}{k}}$ and we can find $t$ such that $P(w_0)=\varepsilon^t$.

We first define $w_0'=\varepsilon^{t+\frac{1}{2}}$.
Moreover, since $T$ is bounded, there exists $M$ such that $$|x_j|\leq M, \qquad \forall\, j=1,\dots ,n .$$

 Now, we can define $\displaystyle \delta=\frac{1}{nM}\sin(\frac{\pi}{k})$. We will show that $w_0'$ and $\delta$ will satisfy the conclusion of the theorem. \\

 For $W=x_1w_1+...+x_nw_n,$ such that $|w_j|<\delta$ for every $j=1,\dots ,n$,  it is easy to note that we have 
$$||W||\leq M(|w_1|+...+|w_n|)< Mn \delta.$$

Thus, we have

\begin{equation}
\displaystyle||W||< Mn \delta=\sin(\frac{\pi}{k}), 
\end{equation}
which yields: 
$$||(W+w'_0)-w_0'||=||W||< \sin(\frac{\pi}{k}).$$

Therefore $W+w'_0$ belongs to the ball $D(w_0',R)$ centered in $w_0'$ and with radius $R=\sin(\frac{\pi}{k})$.
This ball is tangent to the rays $\varepsilon^t$ and $\varepsilon^{t+1}$, therefore for every $z$ in $D(w_0',R)$ we have:
$$\frac{2\pi t}{k} \leq \arg(z)< \frac{2\pi(t+1)}{k}. $$

Hence, in particular:
$$\frac{2\pi t}{k} \leq Arg(w_0'+W)<\frac{2\pi(t+1)}{k},$$
 for any $W=x_1w_1+...+x_nw_n,$ such that $|w_j|<\delta$.

We then obtain that:

$$P(w_0'+\sum_{l=1}^nw_lx_l)=P(w_0),$$ 
which yields:
$$f(x_1,...,x_n)=P(w_0'+\sum_{l=1}^nw_lx_l),\qquad \forall (x_1,...,x_n)\in T.$$

\end{proof}

The theorem above guarantees the existence of a threshold function and we now write the separability conditions under which the complex algorithm becomes convergent.
\begin{definition}
\label{k-sep}
We call the sets $\{A_l\}_{1\leq l\leq k}$, where $A_l \subset \mathbb{C}^n$, $k-$separable, if and only if there exists $W\in\mathbb{C}^n$ and a permutation $\{\alpha_l\}_{1\leq l\leq k}$ of $\{1,\dots, k\}$ such that $P(\langle X,\, W\rangle)=\varepsilon^{\alpha_l}$ for any $X\in A_l$.
\end{definition}

We can now define the MVN complex learning algorithm. Given $k-$separable sets $A_l \subset \mathbb{C}^n$ one can write the following learning rule for the complex perceptron:

\begin{equation}
\label{calg}
W_{k+1}=W_k+C_k(\varepsilon^{q_k}-\varepsilon^{s_k})\overline{X_k},
\end{equation}
where $X_k\in\cup_{l=1}^{l=k} A_l$, $q_k$ is the desired output and $s_k$ is the actual output. One can see that for vectors already organized in their desired sets the algorithm will stop,
This algorithm will find a vector $W$ in a finite number of steps (this vector may not be unique) and in the following theorem we will write and prove the convergence for the Perceptron Theorem in the complex case.
\begin{theorem}\label{Complexthm}
Under the $k-$separability condition the MVN learning algorithm converges after a finite number of steps.
\end{theorem}
\begin{proof}
We will prove that $$\mathcal{O}(k^2)\leq ||W_{k+1}||^2 \leq \mathcal{O}(k),$$ which yields convergence. For the first part we will use the $k-$separability condition. For the algorithm mentioned above:
$$W_{k+1}=W_k+C_k(\varepsilon^{q_k}-\varepsilon^{s_k})\overline{X_k},$$
and we can reorganize all the vectors $X_k$ using the separation hypothesis.
Indeed, we write $$W_{k+1}=W_k+(1-\varepsilon^{s_k-q_k})\overline{X'_k}.$$

\begin{itemize}
\item If $P(\langle X'_k,W_k \rangle) =\varepsilon^0$ we remove the elements for which we have $W_{k+1}=W_k$ and denote the remaining ones by $\overset{\sim}X_k$.
\item If $P(\langle X'_k,W_k \rangle) \neq \varepsilon^0$ we let 
$$\overset{\sim}W_{k+1}=(1-\varepsilon^{s_1-q_1})\overline{\overset{\sim}X_1}+...+(1-\varepsilon^{s_k-q_k})\overline{\overset{\sim}X_k}$$

\end{itemize}
Then, by hypothesis there exists $W$ such that 
\begin{equation} 
P(\langle X', W \rangle)=\varepsilon^0.
\end{equation}

Moreover, it holds that 
$$\displaystyle \langle \overset{\sim}W_{k+1}, W \rangle =\sum_{l=1}^k(1-\varepsilon^{s_l-q_l}) \langle \overline{\overset{\sim}X_l}, W \rangle.$$
Thus, we get

$$\displaystyle \overline{\langle \overset{\sim}W_{k+1}, W \rangle} =\sum_{l=1}^k(1-\varepsilon^{q_l-s_l})\langle \overset{\sim}X_l, \overline{W} \rangle.$$
We have $\displaystyle Arg(\overline{1-\varepsilon^{r_l}})=\frac{\pi}{2}-\frac{\pi r_l}{k}$ and $ 0 < r_l\leq k-2$ where $r_l=s_l-q_l \textbf{ } mod(k)$. We note that $Arg(1-\varepsilon^{r_l})=-\frac{\pi}{2}+\frac{\pi r_l}{k}.$ 
It holds that 

$$-\frac{\pi}{2}+\pi\frac{r_l}{k}\leq Arg((1-\varepsilon^{r_l})\langle  \overset{\sim}X_l, \overline{W}\rangle) \leq -\frac{\pi}{2}+\pi\frac{r_l+2}{k}$$

Therefore, taking the maximum of $r_l$ from the right and its minimum from the left we get 

$$-\frac{\pi}{2}< Arg((1-\varepsilon^{r_l})\langle  \overset{\sim}X_l, \overline{W}\rangle) \leq -\frac{\pi}{2}+\pi\frac{k}{k}=\frac{\pi}{2}.$$
Thus, $\Re \left((1-\varepsilon^{s_l-q_l}) \langle \overline{\overset{\sim}X_l}, W \rangle \right)\geq 0$. Then, setting $m=\underset{l=1,...,k} \min\left(\Re \left((1-\varepsilon^{s_l-q_l}) \langle \overline{\overset{\sim}X_l}, W \rangle \right)\right)$ we obtain 

\begin{equation}\label{i1}
| \langle \overset{\sim}W_{k+1}, W \rangle |\geq k m.
\end{equation}
However, using the Cauchy Schwarz inequality we know also that 

$$|\langle \overset{\sim}W_{k+1}, W \rangle|\leq ||\overset{\sim}W_{k+1}|| \textbf{ } ||W||.$$
In particular, if we combine the previous inequality with \eqref{i1} we obtain 
\begin{equation}
\frac{m^2}{||W||^2}k^2\leq ||\overset{\sim}W_{k+1}||^2.
\end{equation}
For the second part, since $W_{k+1}=W_k+C_k(\varepsilon^{q_k}-\varepsilon^{s_k})\overline{X_k},$
we have $$||W_{k+1}||^2=||W_k||^2+C_k^2 ||\varepsilon^{q_k}-\varepsilon^{s_k}||^2||X_k||^2+2C_k\, \Re[\overline{W_k}\,\overline{X_k}(\varepsilon^{q_k}-\varepsilon^{s_k})]$$
Thus, 
\begin{equation}\label{wk1}
||W_{k+1}||^2=||W_k||^2+C_k^2 ||\varepsilon^{q_k}-\varepsilon^{s_k}||^2||X_k||^2+2C_k \, \Re[\overline{W_k}\,\overline{X_k}(\varepsilon^{q_k}(1-\varepsilon^{s_k-q_k})].
\end{equation}

We note also that 
$$ \frac{2\pi(n-s_k-1)}{n}\leq Arg(\overline{W_k}\,\overline{X_k})\leq \frac{2\pi(n-s_k)}{n}.$$
Then, using trigonometric identities we observe that:

\[
    \begin{split}
      \displaystyle \varepsilon^{q_k}-\varepsilon^{s_k}  &= \varepsilon^{q_k}(1-\varepsilon^{s_k-q_k})  \\
        &=\varepsilon^{q_k}\left(1-\cos\left(\frac{2\pi(s_k-q_k)}{n}\right)-i\sin\left(\frac{2\pi(s_k-q_k)}{n}\right)\right)
         \\
        &=\varepsilon^{q_k}\left(2\sin^2\left(\frac{2\pi(s_k-q_k)}{2n}\right)-2i\sin\left(\frac{2\pi(s_k-q_k)}{2n}\right)\cos\left(\frac{2\pi(s_k-q_k)}{2n}\right)\right)
         \\
        &=2(-i)\varepsilon^{q_k}\sin\left(\frac{2\pi(s_k-q_k)}{2n}\right)\left(\cos\left(\frac{2\pi(s_k-q_k)}{2n}\right)+i\sin\left(\frac{2\pi(s_k-q_k)}{2n}\right)\right)     \\
        &= 2e^{(\frac{\pi(q_k+s_k)}{n}-\frac{\pi}{2})i} \sin\left(\frac{2\pi(s_k-q_k)}{2n}\right) 
           \\
        &=2\left(\cos\left(\frac{\pi(q_k+s_k)}{n}-\frac{\pi}{2}\right)+i\sin\left(\frac{\pi(q_k+s_k)}{n}-\frac{\pi}{2}\right) \right)\sin\left(\frac{2\pi(s_k-q_k)}{2n}\right).
           \\
        &
          \end{split}
   \]
Therefore, we obtain $$\varepsilon^{q_k}(1-\varepsilon^{s_k-q_k})=2\left(\cos\left(\frac{\pi(q_k+s_k)}{n}-\frac{\pi}{2}\right)+i\sin\left(\frac{\pi(q_k+s_k)}{n}-\frac{\pi}{2}\right) \right)\sin\left(\frac{2\pi(s_k-q_k)}{2n}\right).$$
Setting $r=||X_kW_k||$ and $\mu=Arg(\overline{W_k}\,\overline{X_k})$ we have 

\[
    \begin{split}
      \displaystyle \Re(\overline{W_k}\,\overline{X_k}(\varepsilon^{q_k}-\varepsilon^{s_k}))  &= \Re(\overline{W_k}\,\overline{X_k}\varepsilon^{q_k}(1-\varepsilon^{s_k-q_k})) \\
      &\hspace{-1.5cm}=2\Re\left(r e^{\mu i}\left(\cos\left(\frac{\pi(q_k+s_k)}{n}-\frac{\pi}{2}\right)+i\sin\left(\frac{\pi(q_k+s_k)}{n}-\frac{\pi}{2}\right) \right)\sin\left(\frac{2\pi(s_k-q_k)}{2n}\right)\right) \\
        &\hspace{-1.5cm}=2\Re\left(re^{\mu i}e^{\frac{\pi(q_k+s_k)}{n}i}e^{-\frac{\pi}{2}i}\right)\sin\left(\frac{2\pi(s_k-q_k)}{2n}\right)
        \\
        &\hspace{-1.5cm}=2r \,\Re \left(-i e^{(\mu+\frac{\pi(q_k+s_k)}{n})i} \right)\sin(\frac{\pi(s_k-q_k)}{n})
        \\
        &\hspace{-1.5cm}=2r\sin(\mu+\frac{\pi(s_k+q_k)}{n})\sin(\frac{\pi(s_k-q_k)}{n})
          \end{split}
   \]
We note that for $\displaystyle 2\pi-\frac{2\pi(s_k+1)}{n}\leq \mu < 2\pi-\frac{2\pi s_k}{n} $ we have  $$2\pi+2\pi\frac{(q_k-s_k)}{2n}-\frac{2\pi}{n}< \mu+\frac{\pi(s_k+q_k)}{n}< 2\pi+\frac{2\pi(q_k-s_k)}{2n}.$$
Hence, we have $$ \displaystyle \Re(\overline{W_k}\,\overline{X_k}(\varepsilon^{q_k}-\varepsilon^{s_k}))=2r\sin(\mu+\frac{\pi(s_k+q_k)}{n})\sin(\frac{\pi(s_k-q_k)}{n})\leq 0.$$
Then, inserting the previous inequality in~\eqref{wk1} we obtain 
$$||W_{k+1}||^2\leq ||W_k||^2+2C_k^2||X_k||^2\leq ||W_k||^2+M,$$
with $M>0$. 
Hence, by iteration we obtain 
$$||W_{k+1}||^2\leq kM.$$
Therefore, combining both the first and second parts we obtain:
\begin{equation}
\frac{m^2}{||W||^2}k^2 \leq ||W_{k+1}||^2\leq k M,
\end{equation}
which completes the proof of the convergence algorithm in the complex case.
\end{proof}
\begin{remark}
A similar proof was written in~\cite{georgiou}, we are including our own version for completion. An interested reader may also consult the books of I. Aizenberg and N. Aizenberg~\cite{aizenbergtwice,aizenbergson}.
\end{remark}


\section{The convergence of the perceptron algorithm in the bicomplex case}
\setcounter{equation}{0}
\label{BC_perceptron}

In order to consider the bicomplex convergence algorithm we explore first the existence of a bicomplex threshold function, and the definition of separability in this context, using the hyperbolic valued norm as in Section~\ref{BCvectors_norm}, an extension to the norm described in~\cite{alss}.

\subsection{Existence of activation functions}

We have the following definition of the activation function $\mathcal{P}$ in the bicomplex case :
\begin{definition}
For $\mathsf{w}_0=w_{0_1}\mathbf{e}_1+w_{0_2}\mathbf{e}_2$, $\mathsf{w}_\ell=w_{\ell_1}\mathbf{e}_1+w_{\ell_2}\mathbf{e}_2$ and $\mathsf{x_\ell}=x_{\ell_1}\mathbf{e}_1+x_{\ell_2}\mathbf{e}_2$ in $\mathbb{BC}$ with $\ell=1,...,n$. We define the bicomplex activation function by
$$\mathcal{P}(\mathsf w_0+\sum_{\ell=1}^{n}\mathsf w_{\ell}\mathsf x_{\ell}):=P(w_{0_1}+\sum_{\ell=1}^{n}w_{\ell_1}x_{\ell_1})\mathbf{e}_1+P(w_{0_2}+\sum_{\ell=1}^{n}w_{\ell_2}x_{\ell_2})\mathbf{e}_2,$$
where $P$ is the complex activation function given by Definition \ref{ActFunComplex}.
\end{definition}

We can now generalize the notion of threshold function as follows:
\begin{definition}
Let $n\geq 1$ and $T\subset \mathbb{BC}^n$. Then, a complex valued function $f:T\longrightarrow \mathbb{BC}$ is called a $\mathbb{BC}$-threshold function if there exists a weighting vector $\mathsf W=(\mathsf w_0,\mathsf w_1,...,\mathsf w_n)\in \mathbb{BC}^{n+1}$ such that:
\begin{equation}
f(\mathsf x_1,...,\mathsf x_n)=\mathcal{P}(\mathsf w_0+\sum_{\ell=1}^{n}\mathsf w_{\ell}\mathsf x_{\ell}), \quad \forall
(\mathsf x_1,...,\mathsf x_n)\in T.
\end{equation}
\end{definition}

\begin{theorem}
Let $T$ be a bounded domain of $\mathbb{BC}^n$ (i.e: there exist $T_1,T_2\subset \mathbb{C}^n$ which are bounded such that we have $T=T_1\mathbf{e}_1+T_2\mathbf{e}_2).$ Let $f:T\longrightarrow \mathbb{BC}$ a bicomplex threshold function and $(\mathsf w_0,0,...,0)$ a weighting vector of $f(\mathsf x_1,...,\mathsf x_n)$. Then, there exists $\mathsf w'_0\in\mathbb{BC}$ and $\delta>0$ such that $(\mathsf w'_0,\mathsf w_1,...,\mathsf w_n)$ is a weighting vector of $f$ for every $\mathsf w_1,...,\mathsf w_n$ that satisfy $|\mathsf w_l|\preceq \delta\mathbf{e}_1+\delta\mathbf{e}_2$ (i.e: $|\omega_{l_1}|<\delta$ and $|\omega_{l_2}|<\delta$).
\end{theorem}
\begin{proof}
We take $\delta=\min(\delta_1,\delta_2)$ and define $\mathsf w_0'=\varepsilon^{t_1+\frac{1}{2}}\mathbf{e}_1+\varepsilon^{t_2+\frac{1}{2}}\mathbf{e}_2$.

Thus, we have $$\mathsf w_0'=\frac{\varepsilon^{\frac{1}{2}}}{2}((\varepsilon^{t_1}+\varepsilon^{t_2})+(\varepsilon^{t_1}-\varepsilon^{t_2})\mathbf{j}).$$
\end{proof}


\subsection{Bicomplex Perceptron Theorem}

In order to write the bicomplex perceptron algorithm, we first introduce the notion of separability for bicomplex sets.
\begin{definition}
\label{k-sepBC}
Let us consider the sets of $\mathbb{BC}^{n}$ given by:
\begin{equation}
A_l=A_{l_1}\mathbf{e_1}\oplus A_{l_2}\mathbf{e_2},
\end{equation}
where $A_{l_1}$ and $A_{l_2}$ are subsets of $\mathbb{C}^n$. We say that the sets $(A_l)_{1 \leq l\leq k}$ are {\em $k$-separable} if and only if there exists $\mathsf W\in\mathbb{BC}^n$ such that if $\mathsf W=\omega_1\mathbf{e}_1+\omega_2 \mathbf{e}_2$ we have that 

$(A_{l_1})_{1 \leq l\leq k}$ and $(A_{l_2})_{1 \leq l\leq k}$ are $k$-separable with respect to $\omega_1$ and $\omega_2$. In other words, there exist two permutations $(\pi_{l_1})_{1\leq l\leq k}$ and $(\pi_{l_2})_{1 \leq l\leq k}$ such that  
\begin{equation}
P(\langle X_1,\, \omega_1 \rangle)=\varepsilon^{\pi_{l_1}}, \quad \forall X_1\in A_{l_1}
\end{equation}

and \begin{equation}
P(\langle X_2,\, \omega_2\rangle)=\varepsilon^{\pi_{l_2}},\quad \forall X_2\in A_{l_2},
\end{equation}
where $\displaystyle\varepsilon=\exp(\frac{2\pi i}{k})$ and $P$ is the complex activation function used in Section~\ref{complex_Perc}.
\end{definition}

\begin{theorem}
Under the $k-$separability condition as in Definition~\ref{k-sepBC}, the bicomplex perceptron algorithm converges after a finite number of steps given by $n=\max(n_1,n_2)$ where $n_1$ and $n_2$ are the number of steps given by the two respective complex perceptron algorithms from the decomposition.
\end{theorem}
\begin{proof}
We consider the following learning rule in the bicomplex setting:
\begin{equation}\label{LBC}
\mathsf W_{k+1}=\mathsf W_k+C\xi_k \mathsf X_k^*,
\end{equation}

where $C\in\mathbb R$ and $$\xi_k=(\varepsilon^{q_{1,n}}-\varepsilon^{s_{1,n}})\mathbf{e_1}+(\varepsilon^{q_{2,n}}-\varepsilon^{s_{2,n}})\mathbf{e_2}.$$
We should use the representation in $Z=\lambda_1 \mathbf{e_1}+\lambda_2 \mathbf{e_2}$ in order to prove the results then translate everything in terms of $Z=z_1+z_2 j$. We will use the following notations for each term in \eqref{LBC}: 
$$\mathsf W_{k}=W_{k,1}+\mathbf{j}W_{k,2}=w_{k,1}\mathbf{e_1}+w_{k,2}\mathbf{e_2},$$
and $$\mathsf X_k=X_{k,1}+\mathbf{j}X_{k,2}=x_{k,1}\mathbf{e_1}+x_{k,2}\mathbf{e_2}.$$
The conjugate considered in \eqref{LBC} is the one defined by $\mathsf X_k^*=\overline{x_{k,1}}\mathbf{e_1}+\overline{x_{k,2}}\mathbf{e_2},$ and $\rho_l $ in each component of $\xi$ corresponds to a segment on the respective unit disk as in~\cite{georgiou}.
Thus, using the relations of $\mathbf{e}_1$ and $\mathbf{e}_2$ we have 
\begin{equation}
\mathsf W_{k+1}=(w_{k,1}+C(\varepsilon^{q_{1,n}}-\varepsilon^{s_{1,n}})\overline{x_{k,1}})\mathbf{e_1}+(w_{k,2}+C(\varepsilon^{q_{2,n}}-\varepsilon^{s_{2,n}})\overline{x_{k,2}})\mathbf{e_2}
\end{equation}

Hence, we obtain 
\begin{equation}
||\mathsf W_{k+1}||_{\mathbb{D}_+}= ||w_{k,1}+C(\varepsilon^{q_{1,n}}-\varepsilon^{s_{1,n}})\overline{x_{k,1}}||\mathbf{e}_1+||w_{k,2}+C(\varepsilon^{q_{2,n}}-\varepsilon^{s_{2,n}})\overline{x_{k,2}}||\mathbf{e}_2
\end{equation}
It follows that:
\begin{equation}
||\mathsf W_{k+1}||_{\mathbb{D}_+}= ||w_{k+1,1}||\mathbf{e}_1+||w_{k+1,2}||\mathbf{e}_2,
\end{equation}

 with the complex learning rules on each component can be written as:

$$w_{k+1,1}=w_{k,1}+C(\varepsilon^{q_{1,n}}-\varepsilon^{s_{1,n}})\overline{x_{k,1}},$$

and $$w_{k+1,2}=w_{k,2}+C(\varepsilon^{q_{2,n}}-\varepsilon^{s_{2,n}})\overline{x_{k,2}}.$$

Thus, using the complex perceptron convergence theorem proved in Theorem \ref{Complexthm} we know that there exists $w_1$ and $w_2$ such that 

$$\frac{k^2}{||w_1||^2}\leq ||w_{k+1,1}||\leq kM, $$

and $$\frac{k^2}{||w_2||^2}\leq ||w_{k+1,2}||\leq kM. $$ Now we can consider the bicomplex number given by $W=w_1\mathbf{e}_1+w_2\mathbf{e}_2$, we have 
$$||\mathsf W||_{\mathbb{D}_+}=||w_1||\mathbf{e}_1+||w_2||\mathbf{e}_2,$$
and $$ \frac{1}{||\mathsf W||_{\mathbb{D}_+}}=\frac{1}{||w_1||}\mathbf{e}_1+\frac{1}{||w_2||_{\mathbb{D}}}\mathbf{e}_2.$$
Hence, we deduce that 
\begin{equation}
\frac{k^2}{||\mathsf W||_{\mathbb{D}_+}}\preceq ||\mathsf W_{k+1}||_{\mathbb{D}_+}\preceq kM \mathbf{e}_{1}+kM\mathbf{e}_2=kM
\end{equation}

This ends the proof of the convergence of the bicomplex perceptron algorithm.
\end{proof}

We will now return to the bicomplex setting expressed in terms of the complex units $\i,\j$ and re-write the algorithm in this case, for ease of computation and to highlight the bicomplex structure in a way that will be used in other generalizations in the future.

We have
\begin{eqnarray*}
&\mathsf W_{k+1}= w_{k+1,1}\mathbf{e_1}+w_{k+1,2}\mathbf{e_2} =\\
&(w_{k,1}+C(\varepsilon^{q_{1,n}}-\varepsilon^{s_{1,n}})\overline{x_{k,1}})\mathbf{e_1}+(w_{k,2}+C(\varepsilon^{q_{2,n}}-\varepsilon^{s_{2,n}})\overline{x_{k,2}})\mathbf{e_2}\\
&=W_{k+1,1}+\j W_{k+1,2}.
\end{eqnarray*}

\begin{corollary}
The bicomplex algorithm can be re-written as:
\begin{equation}
   \displaystyle \mathsf W_{k+1} = \mathsf W_{k}+(1+\mathbf{ij})\frac{C}{2}\left((\varepsilon^{q_{1,n}}-\varepsilon^{s_{1,n}})\mathsf X_k^* +(\varepsilon^{q_{2,n}}-\varepsilon^{s_{2,n}})\overline{\mathsf X_k} \,\right). 
  \end{equation}
\end{corollary}

\begin{proof}
We have:

\begin{eqnarray*}
W_{k+1,1}=W_{k,1}+\frac{C}{2} [(\varepsilon^{q_{1,n}}-\varepsilon^{s_{1,n}})\overline{(X_{k,1}-\mathbf{i}X_{k,2})} +(\varepsilon^{q_{2,n}}-\varepsilon^{s_{2,n}})\overline{(X_{k,1}+iX_{k,2})}]\\
W_{k+1,2}=W_{k,2}+\frac{\mathbf{i}C}{2} [(\varepsilon^{q_{1,n}}-\varepsilon^{s_{1,n}})\overline{(X_{k,1}-\mathbf{i}X_{k,2})} -(\varepsilon^{q_{2,n}}-\varepsilon^{s_{2,n}})\overline{(X_{k,1}+iX_{k,2})}]
\end{eqnarray*}
We recall that $\mathsf X_k=X_{k,1}+\mathbf{j}X_{k,2}$ and use the following bicomplex conjugates $$\mathsf X_k^*=\overline{X_{k,1}}-\mathbf{j}\overline{X_{k,2}}, \quad \overline{\mathsf X_{k}}=\overline{X_{k,1}}+\mathbf{j}\overline{X_{k,2}}.$$ Then, we easily observe that $$\displaystyle \overline{X_{k,1}}=\frac{1}{2}(\mathsf X_k^*+\overline{\mathsf X_k}), \quad \overline{X_{k,2}}=\frac{\mathbf{j}}{2}(\mathsf X_k^*-\overline{\mathsf X_k}).$$
Thus, we have $$\overline{X_{k,1}-\mathbf{i}X_{k,2}}=\frac{1}{2}\left((1+\mathbf{ij})\mathsf X_k^*+(1-\mathbf{ij})\overline{\mathsf X_k}\right), $$
and 
$$\overline{X_{k,1}+\mathbf{i}X_{k,2}}=\frac{1}{2}\left((1-\mathbf{ij})\mathsf X_k^*+(1+\mathbf{ij})\overline{\mathsf X_k}\right).$$
Hence, we obtain \[
    \begin{split}
      \displaystyle W_{k+1,1} &=  W_{k,1}+\frac{C}{4}\left[(\varepsilon^{q_{1,n}}-\varepsilon^{s_{1,n}})\left((1+\mathbf{ij})\mathsf X_k^*+(1-\mathbf{ij})\overline{\mathsf X_k}\right)+(\varepsilon^{q_{2,n}}-\varepsilon^{s_{2,n}})\left((1-\mathbf{ij})\mathsf X_k^*+(1+\mathbf{ij})\overline{\mathsf X_k}\right)\right], \\
      &
          \end{split}
   \]
   
   and 
   
   \[
    \begin{split}
      \displaystyle W_{k+1,2} &= W_{k,2}+\frac{C}{4}\left[(\varepsilon^{q_{1,n}}-\varepsilon^{s_{1,n}})\left((\mathbf{i}-\mathbf{j})\mathsf X_k^*+(\mathbf{j}+\mathbf{i})\overline{\mathsf X_k}\right)+(\varepsilon^{q_{2,n}}-\varepsilon^{s_{2,n}})\left((\mathbf{j}+\mathbf{i})\mathsf X_k^*+(\mathbf{i}-\mathbf{j})\overline{\mathsf X_k}\right) \right]. \\
      &
          \end{split}
   \]
  Finally, we obtain 
  \begin{equation}
   \displaystyle \mathsf W_{k+1} = \mathsf W_{k}+(1+\mathbf{ij})\frac{C}{2}\left((\varepsilon^{q_{1,n}}-\varepsilon^{s_{1,n}})\mathsf X_k^* +(\varepsilon^{q_{2,n}}-\varepsilon^{s_{2,n}})\overline{\mathsf X_k} \,\right). 
  \end{equation}
This completes the convergence algorithm in the standard bicomplex form.
\end{proof}


\section{Conclusions and Future Work}

The authors are working to establish other perceptron theorems in different hypercomplex settings, as well as different algorithms in the bicomplex case.
Interesting future work will include the quaternionic case, the hyperbolic case, as well as higher dimension hypercomplex algebras, commutative or not. 
These algorithms can simplify and reduce errors and be quite useful in their implementation.
Hypercomplex valued neural networks have the ability of accumulating several complex variables into a single variable theory that can ease calculations and improve convergence.

\bibliographystyle{plain}

\begin{thebibliography}{10}

\bibitem{agmon}
Agmon,~S., "The relaxation method for linear inequalities",  Canad. J. Math.,  Vol. 40 (1954), pp. 382--392.


\bibitem{abbn} Ahmadi, S., Beyhaghi, H., Blum, A., Naggita, K. "The strategic perceptron". In Proceedings of the 22nd ACM Conference on Economics and Computation (2021) pp. 6-25.

\bibitem{aizenbergtwice} Aizenberg,~I.N., Aizenberg,~N.N, Vandewalle,~J. {\em Multi-Valued and Universal Binary Neurons, Theory, Learning, and Applications}, Kluwer Academic Publishers, 2000

\bibitem{aizenbergson} Aizenberg,~I.N. {\em Complex Valued Neural Networks with Multi-Valued Neurons}, Studies in Computational Intelligence, Vol. 353, Springer, (2011)


\bibitem{AIPK1973} Aizenberg, N. N., Ivas’kiv, Y.L., Pospelov, D. A., Khudyakov, G. F., “Multivalued threshold functions,”Cybernetics,vol.9 (1973) pp. 61–77.

\bibitem{AZivko}  Aizenberg, N. N.,  Zivko Tosic, “A generalization of the threshold functions,” Publikacije Elektrotehnickog fakulteta. Serija Matematika i fizika, no. 381/409, pp. 97–99, 1972.

 \bibitem{AG} Alfsmann,~D., G\"ockler,~H.G., "Design of Hypercomplex Allpass-Based Paraunitary Filter Banks applying Reduced Biquaternions", Euroconn 2005, Serbia \& Montenegro, Belgrade, November 22-24, 2005

\bibitem{AGES} Alfsmann,~D., G\"ockler,~H.G., Sangwine,~S.J., and Ell,~T.,  "Hypercomplex Algebras in Digital Signal Processing: Benefits and Drawbacks" in 15th European Signal Processing Conference (EUSIPCO 2007), Poznan, Poland, September 3-7, 2007


\bibitem{alss} Alpay,~D., Luna-Elizarraras,~E., Shapiro,~M., Struppa,~D.,
  {\em Basics of Functional Analysis with Bicomplex Scalars, and Bicomplex Schur Analysis} SpringerBriefs in Mathematics, Springer, Cham, 2014.


\bibitem{BMPU} Benvenuto, N., Marchesi, M., Piazza, F., Uncini, A., “Non linear satellite radio links equalized using blind neural networks,” pp. 1521 – 1524 vol.3 (1991), 05.

\bibitem{block1962perceptron} Block,~H.D., "The perceptron: A model for brain functioning {I}", Reviews of Modern Physics., 34(1) (1962): 123.

\bibitem{BS2008}
Buchholz, S., Sommer, G. "On Clifford neurons and Clifford multi-layer perceptrons". Neural Networks, 21(7) (2008), pp.925-935.

\bibitem{CSVV} Colombo,~F., Sabadini,~I., Struppa,~D.C., Vajiac,~A., Vajiac~M.B., "Singularities of functions of one and several bicomplex variables",  Ark. Mat., Vol. 49, no. 2 (2011), pp. 277-294.
  
    
\bibitem{DSVV} De Bie,~H., Struppa,~D., Vajiac,~A., Vajiac,~M., "The Cauchy‐Kowalewski product for bicomplex holomorphic functions", Math. Nachr. Vol. 285, Issue 10 (2012), 1230--1242

 
\bibitem{duda} Duda,~R.O.,  Hart,~P.E, {\em Pattern Classification and Scene Analysis}, Wiley; (1st edition 1973)


\bibitem{georgiou} Georgiou,~G.M. "The multivalued and continuous perceptron", The World congress in Neural Networks, Volume IV (1993) pp. 679-683, Portland, OR.

\bibitem{haykin}  Haykin,~S {\em Neural Networks and Learning Machines} (3rd Edition), Pearson, (2009)

\bibitem{hebb1949organisation} Hebb,~D.O.,  {\em The organisation of behaviour: a neuropsychological theory}, Science Editions New York, 1949.


\bibitem{Hirose-Yoshida} Hirose, A., Yoshida, S. “Generalization characteristics of complex- valued feedforward neural networks in relation to signal coherence,” IEEE Transactions on Neural Networks and Learning Systems, vol. 23 (2012) pp. 541–551.

\bibitem{Huang} Huang, R. C, Chen, M.S, “Adaptive equalization using complex- valued multilayered neural network based on the extended Kalman filter,” vol. 1 (200) pp. 519 – 524 vol.1, 02.


\bibitem{LWYZ}
Liu, W., Gao, P., Wang, Y., Yu, W., Zhang, M. "A unitary weights based one-iteration quantum perceptron algorithm for non-ideal training sets." IEEE Access, 7 (2019), pp.36854-36865.



\bibitem{bcbook}  Luna-Elizarraras,~E., Shapiro,~M., Struppa,~D., Vajiac ,~A. {\em Bicomplex Holomorphic Functions. The Algebra, Geometry and Analysis of Bicomplex Numbers}, Birkh\"auser/Springer, Cham, 2015.
  
  \bibitem{mcpitts} McCullogh,~W.S., and Pitts,~W., "A logical calculus of the ideas immanent in nervous activity",
The bulletin of mathematical biophysics, Vol 5 (1943), pp. 115--133.

\bibitem{minskypapert} Minsky,~M.L., Papert,~S.A.,  {\em {Perceptrons. expanded ed}},  London: MIT Press, expanded ed. edition, 1988.

\bibitem{MR62787} Motzkin,~T.S., and Schoenberg,~I.J., "The relaxation method for linear inequalities",  Canad. J. Math., Vol. 6 (1954), pp. 393--404.

\bibitem{nilsson1965learning} Nilsson,~N.J. {\em Learning machines. Foundations of trainable pattern-classifying systems}, Mc Graw Hill Book Company, New York, 1965.

\bibitem{zbMATH03189712} Novikoff,~A., "On convergence proofs for perceptrons", Proc. Sympos. math. Theor. Automata, New York, April 24-26, 1962, 615-622 (1963).

 \bibitem{Price} Price,~G.B., {\em An Introduction to Multicomplex Spaces and Functions}, Monographs and Textbooks in Pure and Applied Mathematics, \textbf{140}, Marcel Dekker, Inc., New York, 1991.
  

\bibitem{RS2014}  Reichert, D.P., Serre T. “Neuronal synchrony in complex-valued deep networks,” CoRR, vol. abs/1312.6115, 2014.


\bibitem{MR0135635}
Rosenblatt, F.
\newblock {\em Principles of neurodynamics. {P}erceptrons and the theory of
  brain mechanisms}.
\newblock Spartan Books, Washington, D.C., 1962.



\bibitem{SSP} Schuld, M., Sinayskiy, I. and Petruccione, F. Simulating a perceptron on a quantum computer. Physics Letters A, 379(7) (2015), pp.660-663.



\bibitem{Segre} Segre,~C., "Le rappresentazioni reali delle forme complesse e gli enti iperalgebrici", Math. Ann., Vol. 40 (1892), pp. 413--467.


\bibitem{singleton1962test} Singleton,~R.C., "A test for linear separability as applied to self-organizing
  machines", Technical report, Stanford Research Institute, Menlo Park, CA, (1962).


\bibitem{sobczyk} Sobczyk, G., "The Hyperbolic Number Plane",
  Coll. Math. Jour., Vol. 26. No. 4 (1995), pp. 268--280.
 
\bibitem{Sol2002} Solazzi, M., Uncini, A., Di Claudio, E., Parisi, R., “Complex discriminative learning bayesian neural equalizer,” Signal Processing, vol. 81 (2002) pp. 2493–2502, 10. 

 
 \bibitem{TM}Taud, H., Mas, J.F. "Multilayer perceptron (MLP)". In Geomatic Approaches for Modeling Land Change Scenarios (2018), pp. 451-455. Springer, Cham.



\bibitem{mltcplx} Vajiac,~A. and Vajiac,~M., "Multicomplex hyperfunctions", Complex Variables and Elliptic Equations, Vol. 56, Issue 12, pp. 1-13 (2011)

\end{thebibliography}

\end{document}